\documentclass[11pt,a4paper]{article}

\usepackage[T1]{fontenc}
\usepackage[utf8]{inputenc}
\usepackage{lmodern}
\usepackage{microtype}
\usepackage[a4paper,margin=2.25cm]{geometry}
\usepackage{amsmath,amssymb}
\usepackage{booktabs}
\usepackage{graphicx}
\usepackage{subcaption}
\usepackage{tikz}
\usetikzlibrary{arrows.meta,positioning,fit}
\usepackage{natbib}
\usepackage{hyperref}
\usepackage{xcolor}
\usepackage{enumitem}
\usepackage{siunitx}
\usepackage{array}
\usepackage{authblk}
\usepackage{titlesec}
\usepackage{fancyhdr}

\hypersetup{
  colorlinks=true,
  linkcolor=black,
  citecolor=black,
  urlcolor=blue,
  pdftitle={Behavioral Convergence Without Representational Convergence: Persistent Training-History Dependence in Neural Networks},
  pdfauthor={Ertugrul Mutlu},
  pdfsubject={Persistent training-history dependence in neural networks},
  pdfkeywords={representation similarity, CKA, path dependence, sequential learning, hysteresis}
}
\setlist[itemize]{leftmargin=1.5em,itemsep=0.15em,topsep=0.2em}
\setlist[enumerate]{leftmargin=1.7em,itemsep=0.15em,topsep=0.2em}
\titleformat{\section}{\large\bfseries}{\thesection}{0.6em}{}
\titleformat{\subsection}{\normalsize\bfseries}{\thesubsection}{0.6em}{}
\newcommand{\hrepr}{H_{\mathrm{repr}}}
\newcommand{\sabc}{S_{ABC}}
\newcommand{\sbac}{S_{BAC}}

\title{\textbf{Behavioral Convergence Without Representational Convergence:}\\Persistent Training-History Dependence in Neural Networks}
\author{Ertu\u{g}rul Mutlu}
\affil{Aachen, Germany}
\date{September 2026}

\begin{document}
\maketitle

\begin{abstract}
Neural networks trained toward the same final objective can reach similar predictive performance while retaining internal representations shaped by earlier training history. We study this effect using controlled sequential-training experiments in which paired convolutional networks start from identical weights, experience reversed task orders, and then receive the same deterministic common-relaxation distribution. Across 20 paired MNIST runs, 16 satisfy a predeclared behavioral-matching criterion, yet their matched representations retain a mean history score of 0.139 (95\% bootstrap CI: 0.127--0.153) and approximately 3.1\% prediction disagreement. Extending common relaxation to 50,000 optimizer updates does not erase the measured difference: across five paired seeds, the representation-history score remains 0.190 (95\% bootstrap CI: 0.161--0.219) at the end of the measured horizon while the mean accuracy gap is only 0.18 percentage points. Fresh linear probes show that, with sufficient labeled data, the two histories retain practically equivalent linearly accessible class information. A same-label rotated-MNIST control reproduces the effect: all five paired seeds reach behavioral matching while retaining a mean representation-history score of 0.162. Finally, a matched-learning-rate ReLU--LeakyReLU control reduces the 50,000-update representation residue by 0.040 on average in all five paired seeds, providing directional evidence that activation-mediated plasticity contributes to the persistence of training-history effects. These results provide protocol-scoped evidence that behavioral convergence need not imply representational convergence and that optimization history can leave measurable internal traces after prolonged common training.
\end{abstract}


\section{Introduction}

Neural networks are usually compared through observable behavior. If two models obtain nearly identical accuracy and loss on the same distribution, they are often treated as functionally interchangeable. Their internal states, however, need not be interchangeable. Neural-network training is a sequential optimization process, and the solution reached after training may depend on the trajectory by which the model arrived there.

This motivates a basic question: \emph{if two neural networks experience different training histories and are subsequently trained for a long time on exactly the same final distribution, do their internal representations converge once their behavior becomes similar?}

A direct comparison of networks trained in opposite task orders does not isolate this question cleanly. Sequential training is well known to produce catastrophic forgetting and interference, so the task presented most recently can dominate final performance \citep{goodfellow2013catastrophic,parisi2019continual}. A model trained on $A$ and then $B$ may therefore differ from a model trained on $B$ and then $A$ simply because their final tasks differ.

We address this confound with a \emph{common-relaxation} protocol. Starting from identical parameters, one network receives
\[
\sabc: A \rightarrow B \rightarrow C,
\]
whereas its paired counterpart receives
\[
\sbac: B \rightarrow A \rightarrow C.
\]
The two histories differ only in the order of the first two phases. During $C$, both networks receive the same final distribution, deterministic batch sequence, optimizer configuration, and number of optimizer updates. The common phase is intended to reduce the direct behavioral consequences of the preceding task order.

We then ask whether behavioral similarity during $C$ is accompanied by representational similarity. Representations are compared using centered kernel alignment (CKA), a representation-similarity measure designed to compare learned features across neural networks \citep{kornblith2019similarity}. Our declared representation-history metric is
\[
\hrepr = 1-\frac{\mathrm{CKA}_{\mathrm{conv2}}+\mathrm{CKA}_{\mathrm{fc1}}}{2}.
\]
A value of zero indicates perfect CKA agreement at the two declared layers; larger values indicate greater representational divergence.

The main experiment contains 20 paired seeds. Sixteen of the twenty pairs reach a predeclared behavioral-matching criterion during common relaxation. Nevertheless, their matched representations retain a mean $\hrepr$ of approximately 0.139. This difference is not simply a transient immediately after the task-order switch. A separate five-paired-seed experiment extends common relaxation from 10,000 to 50,000 updates, where $\hrepr$ remains substantial.

We perform three additional sets of controls. First, fresh linear classifiers are trained on frozen representations from the common 10,000-update endpoint. With 500 labeled examples per class, the resulting classifiers are practically equivalent within a predeclared $\pm0.5$ percentage-point margin, indicating that representational difference does not imply that one representation has simply lost the class information required for the task. Second, we repeat the sequential protocol with rotated MNIST, where both histories contain the same labels but different input orientations. The representational effect persists. Third, we replace ReLU with LeakyReLU under a matched learning-rate protocol. The residual history score decreases in all five paired seeds, suggesting that activation-mediated plasticity may influence persistence.

Our contribution is deliberately narrower than a universal claim of neural-network hysteresis. We demonstrate persistent training-history dependence over a finite measured optimization horizon under carefully controlled protocols. The central empirical observation is that behavior can substantially converge while internal representations remain measurably history dependent.

\paragraph{Contributions.} This work provides: (i) a paired common-relaxation protocol that controls initialization and final training conditions while reversing earlier task order; (ii) a 20-paired-seed demonstration of behavioral matching without representational matching; (iii) long-horizon and same-label controls showing persistence beyond the original class-split setting; and (iv) functional and mechanistic probes using fresh linear readouts and a matched ReLU--LeakyReLU comparison.

\begin{figure}[t]
\centering
\begin{tikzpicture}[node distance=0.8cm and 0.9cm,>=Latex,every node/.style={font=\small}]
  \node[draw,rounded corners,minimum width=1.15cm,minimum height=0.7cm] (init1) {same $W_0$};
  \node[draw,rounded corners,right=of init1] (a1) {$A$};
  \node[draw,rounded corners,right=of a1] (b1) {$B$};
  \node[draw,rounded corners,right=of b1,minimum width=1.4cm] (c1) {common $C$};
  \node[right=of c1] (out1) {$\sabc$};
  \draw[->] (init1)--(a1); \draw[->] (a1)--(b1); \draw[->] (b1)--(c1); \draw[->] (c1)--(out1);

  \node[draw,rounded corners,below=0.7cm of init1,minimum width=1.15cm,minimum height=0.7cm] (init2) {same $W_0$};
  \node[draw,rounded corners,right=of init2] (b2) {$B$};
  \node[draw,rounded corners,right=of b2] (a2) {$A$};
  \node[draw,rounded corners,right=of a2,minimum width=1.4cm] (c2) {common $C$};
  \node[right=of c2] (out2) {$\sbac$};
  \draw[->] (init2)--(b2); \draw[->] (b2)--(a2); \draw[->] (a2)--(c2); \draw[->] (c2)--(out2);

  \node[draw,dashed,fit=(c1)(c2),inner sep=0.18cm,label=above:{identical final training stream}] {};
  \node[below=0.45cm of c2,align=center,font=\footnotesize] {Compare accuracy, predictions, CKA, and $\hrepr$\\through common-relaxation updates};
\end{tikzpicture}
\caption{Common-relaxation design. Paired networks start from the same initialization, receive reversed histories $A\!\to\!B$ and $B\!\to\!A$, then receive the same deterministic final training stream $C$.}
\label{fig:protocol}
\end{figure}
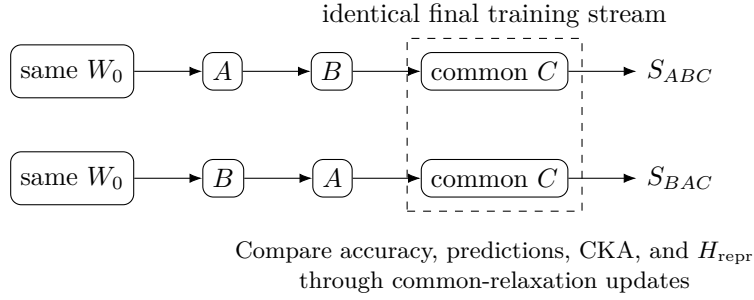

\section{Related Work}

\subsection{Continual learning and catastrophic forgetting}
Sequential neural-network training is strongly associated with catastrophic forgetting: optimization on new data can interfere with parameters and representations learned from earlier data. \citet{goodfellow2013catastrophic} empirically investigated this behavior under gradient-based training, while \citet{parisi2019continual} review catastrophic forgetting as a central challenge in continual and lifelong learning.

Our goal differs from the usual continual-learning objective. We do not propose a method for retaining performance on old tasks. Instead, catastrophic forgetting is a confound that motivates the common-relaxation design. By following different histories with the same $C$ distribution, we ask whether history remains measurable after the most obvious order-dependent behavioral effects have largely disappeared.

\subsection{Comparing neural representations}
Representation-similarity methods make it possible to compare internal computations even when models have similar external performance. SVCCA was introduced by \citet{raghu2017svcca} to compare neural representations across layers and stages of training. \citet{kornblith2019similarity} subsequently developed and analyzed CKA, demonstrating its usefulness for comparing representations across separately trained networks.

We use linear CKA because the question of interest is explicitly representational rather than purely parametric. Weight vectors can differ substantially even when networks implement closely related functions due to parameter symmetries and alternative low-loss solutions. Our primary metric therefore combines CKA at Conv2 and FC1 rather than treating raw Euclidean parameter distance as the principal evidence.

\subsection{Loss landscapes and multiple solutions}
Work on neural-network loss landscapes has shown that apparently different minima can often be connected by low-loss paths \citep{draxler2018barriers,garipov2018loss}. These findings emphasize that parameter-space separation should not automatically be interpreted as functional disconnection. Our question is complementary: even if multiple low-loss solutions are functionally similar or connected, does a model's optimization history leave a measurable signature in its learned representations?

\subsection{Rectifier activations and plasticity}
Rectified activations have shaped modern neural-network optimization. Leaky and parametric variants introduce non-zero gradients in regions where ordinary ReLU has zero derivative \citep{maas2013rectifier,he2015rectifiers}. The related dying-ReLU phenomenon describes units that become inactive over the relevant input domain \citep{lu2019dying}. This motivates our ReLU--LeakyReLU control. We do not assume that inactive ReLU units are the sole source of training-history dependence; instead, the comparison tests whether modifying this aspect of optimization changes the measured residue.

\section{Methods}

\subsection{Dataset and preprocessing}
All experiments use MNIST. Images are transformed to tensors and normalized with mean 0.1307 and standard deviation 0.3081. No data augmentation is applied. Training uses batch size 128, zero data-loader workers, seeded data-loader generators, and requested deterministic PyTorch algorithms.

For the primary class-split protocol,
\[
A=\{0,1,2,3,4\}, \qquad B=\{5,6,7,8,9\}.
\]
Each history contains two sequential ten-epoch phases. Thus $\sabc$ learns $A$ for ten epochs followed by $B$ for ten epochs, whereas $\sbac$ learns $B$ followed by $A$.

The common distribution $C$ is a deterministic balanced MNIST training subset containing 5,000 examples per digit. Its iterator restarts reproducibly, allowing common relaxation to be specified directly by optimizer-update count rather than epochs.

\subsection{Network architecture}
The experiments use a compact convolutional network with no normalization layers. The model contains a $3\times3$ convolution from 1 to 32 channels, max pooling, a $3\times3$ convolution from 32 to 64 channels, a second max-pooling operation, a fully connected layer from $64\times7\times7$ inputs to 128 hidden units, and a ten-class output layer. Padding preserves spatial size across the convolutions.

In the default condition, ReLU follows Conv1, Conv2, and FC1. The mechanism-control condition replaces these activations with LeakyReLU with negative slope 0.01 while preserving all other architecture choices. The implementation exposes post-activation Conv1, Conv2, FC1, and logit representations for analysis.

\subsection{Optimization and common relaxation}
The main class-split experiments use SGD with learning rate 0.05, momentum 0.9, zero weight decay, and batch size 128. An initial pilot compared preserving versus resetting optimizer state at the beginning of $C$. The selected main protocol resets optimizer state immediately before common relaxation; model parameters are unchanged by the reset.

The main experiment evaluates common relaxation at
\[
0,100,500,1000,2500,5000,10000
\]
optimizer updates. The 20 paired seeds are 101, 202, 303, 404, 505, 606, 707, 808, 909, 1010, 1111, 1212, 1313, 1414, 1515, 1616, 1717, 1818, 1919, and 2020. Within each seed, $\sabc$ and $\sbac$ share the same initialization and protocol apart from task order and run identity.

\subsection{Behavioral matching}
For each pair, we select the earliest common-relaxation checkpoint satisfying both
\[
|\mathrm{Acc}_{\sabc}-\mathrm{Acc}_{\sbac}|\leq0.002
\]
and
\[
\min(\mathrm{Acc}_{\sabc},\mathrm{Acc}_{\sbac})\geq0.97.
\]
Thus behavioral matching requires a full-test accuracy gap of at most 0.2 percentage points while both models achieve at least 97\% accuracy. If no checkpoint satisfies the criterion, the seed is marked unmatched.

\subsection{Representational and functional metrics}
Representation comparisons use a deterministic class-balanced test probe containing 200 examples per class. For convolutional layers, activations are reduced by global average pooling; FC1 activations are used directly. Feature-space linear CKA is computed after centering each feature dimension.

The primary representation-history score is
\[
\hrepr = 1-\frac{\mathrm{CKA}_{\mathrm{conv2}}+\mathrm{CKA}_{\mathrm{fc1}}}{2}.
\]
Conv1 and logits are excluded from this declared primary score. Secondary functional metrics include prediction disagreement, Jensen--Shannon divergence between predictive distributions, accuracy gap, and loss difference. Normalized parameter distance and activation-health quantities are retained as secondary diagnostics.

\subsection{Multi-seed aggregation and uncertainty}
All comparisons preserve the paired design; $\sabc$ and $\sbac$ are never treated as independent samples. Mean uncertainty is estimated using 10,000 paired-seed bootstrap resamples with bootstrap seed 12345. Paired $t$ tests, Wilcoxon signed-rank tests, sign counts, and paired Cohen's $d_z$ are reported where relevant.

\subsection{Long-horizon relaxation}
A fresh five-paired-seed experiment uses seeds 101, 202, 303, 404, and 505 and extends common relaxation to 50,000 updates. The selected ReLU/reset protocol is otherwise unchanged. Additional checkpoints are evaluated at 25,000 and 50,000 updates. The interpretation is explicitly finite-horizon: persistence at 50,000 updates is not treated as mathematical permanence.

\subsection{Frozen linear probes}
To test whether representational difference corresponds to different accessibility of class information, we train fresh linear classifiers on frozen FC1 representations from the \emph{final 10,000-update checkpoints} of all 20 main model pairs. This endpoint is fixed across seeds and is not selected based on probe performance.

Backbones are switched to evaluation mode, frozen, and checked for exact equality before and after probe training. Paired SABC and SBAC linear heads begin from exactly identical tensors and receive identical balanced examples, labels, deterministic minibatch permutations, and optimization settings.

The probes use SGD for 20 epochs with learning rate 0.01, zero momentum, zero weight decay, and batch size 128. The default fresh-head seed is 777. Testing uses 200 examples per class. Training budgets of 25, 50, 100, and 500 examples per class are evaluated. For the 50-example condition, robustness is additionally assessed with probe seeds 777, 1777, 2777, 3777, and 4777 while retaining the 20 model seeds as the independent statistical units. At 500 examples per class, an exploratory TOST equivalence analysis uses a $\pm0.5$ percentage-point margin.

\subsection{LeakyReLU mechanism control}
The initial LeakyReLU protocol at learning rate 0.05 exhibited reproducible numerical instability in several seed/order combinations. Rather than compare activation functions under different optimization rates, we performed a descending stability search and reran both ReLU and LeakyReLU at the same selected rate. Learning rate 0.04 was the highest tested candidate that remained finite across all ten short traces (five seeds $\times$ two orders); full long-horizon runs were subsequently audited for non-finite values.

The final matched comparison uses learning rate 0.04, momentum 0.9, zero weight decay, optimizer reset, and 50,000 common-relaxation updates. The predeclared primary endpoint is
\[
\Delta \hrepr=\hrepr^{\mathrm{LeakyReLU}}-\hrepr^{\mathrm{ReLU}}
\]
at 50,000 updates. Negative values indicate lower history residue under LeakyReLU.

\subsection{Same-label rotated-MNIST control}
The class-split experiment changes both input distribution and active label set across $A$ and $B$. To test whether disjoint labels are necessary, we use a rotated-MNIST protocol. Both domains contain all ten digit labels: domain $A$ contains MNIST at $0^\circ$, while domain $B$ contains a deterministic $90^\circ$ rotation. The common distribution is balanced across digits and both rotation domains. Architecture, reset policy, optimizer schedule, checkpoint schedule, and behavioral-matching criterion otherwise follow the main protocol. Five paired seeds are evaluated.

\section{Results}

\subsection{Behavioral matching does not eliminate representational history}
In the 20-paired-seed main experiment, 16 of 20 seeds satisfy the predeclared performance-matching criterion within the first 10,000 common-relaxation updates. At those matched endpoints, mean
\[
\hrepr=0.1387,
\]
with 95\% bootstrap CI $[0.1267,0.1527]$. The score ranges from 0.103 to 0.212 across matched seeds. Mean prediction disagreement remains 3.06\% (95\% bootstrap CI: 2.75--3.40\%). Thus, matching aggregate accuracy does not imply convergence to the same internal representation or even identical per-example predictions.

At the common 10,000-update endpoint across all 20 seeds, mean SABC accuracy is 98.603\% and mean SBAC accuracy is 98.483\%. The mean absolute accuracy gap is 0.252 percentage points and prediction disagreement is 2.74\%. Despite this close behavioral performance, mean $\hrepr=0.1531$ with 95\% bootstrap CI $[0.1439,0.1635]$.

\begin{figure}[t]
\centering
\begin{subfigure}{0.49\textwidth}
  \includegraphics[width=\linewidth]{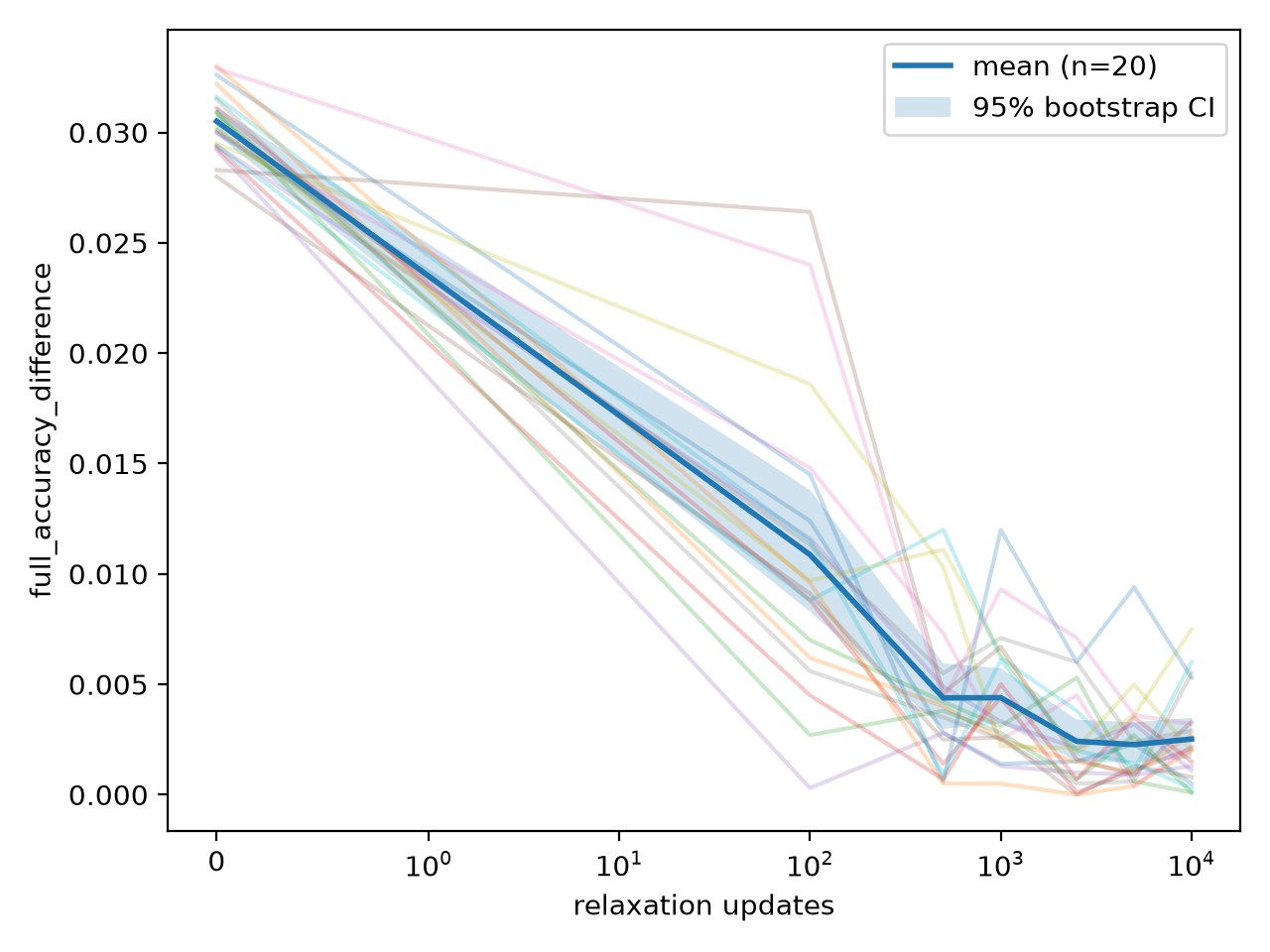}
  \caption{Absolute full-test accuracy gap.}
\end{subfigure}\hfill
\begin{subfigure}{0.49\textwidth}
  \includegraphics[width=\linewidth]{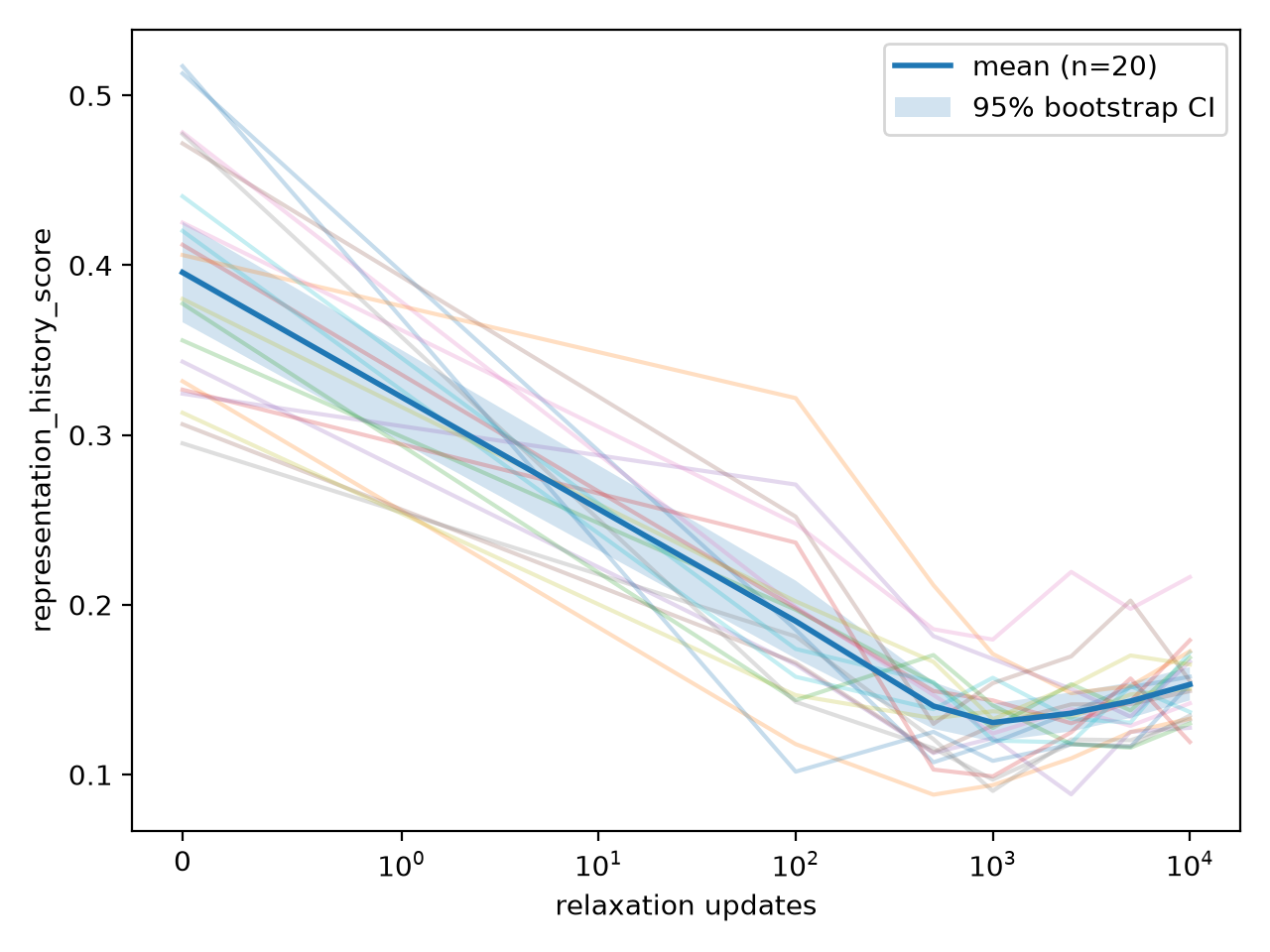}
  \caption{Representation-history score.}
\end{subfigure}
\caption{Main 20-paired-seed common-relaxation experiment. Behavioral differences contract rapidly, while the declared Conv2/FC1 representation-history score remains clearly above zero at the end of the 10,000-update horizon. Thin curves show individual seeds; the heavy curve and band show the aggregate mean and bootstrap interval.}
\label{fig:main}
\end{figure}

\subsection{Fifty thousand common updates do not erase the measured difference}
The long-horizon experiment extends the same ReLU/reset protocol to 50,000 updates across five paired seeds. Mean history scores are 0.1532 at 10,000 updates, 0.1625 at 25,000, and 0.1902 at 50,000. The final bootstrap 95\% CI is $[0.1611,0.2193]$.

At 50,000 updates, mean SABC and SBAC accuracies are 98.480\% and 98.316\%, respectively. The mean absolute accuracy gap is 0.180 percentage points and prediction disagreement is 2.93\%.

The mean change in $\hrepr$ from 10,000 to 50,000 updates is $+0.0369$. Its bootstrap interval is positive, but the paired $t$ test is borderline ($p=0.060$) and the Wilcoxon test gives $p=0.125$. We therefore do not claim systematic monotonic growth. The supported conclusion is narrower: the measured representation residue does not decay toward zero over the observed 50,000-update horizon.

\begin{figure}[t]
\centering
\begin{subfigure}{0.49\textwidth}
  \includegraphics[width=\linewidth]{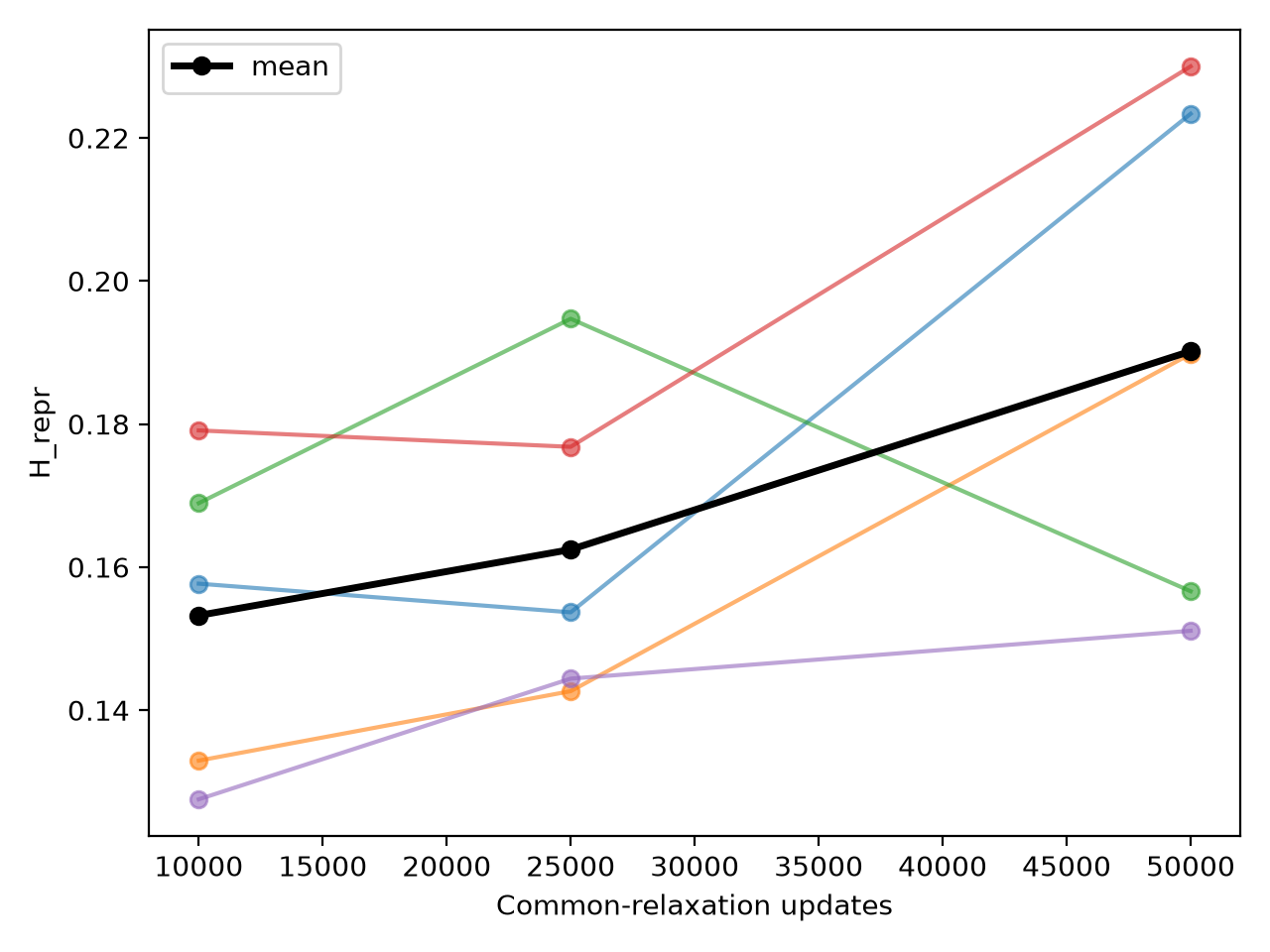}
  \caption{$\hrepr$ from 10k to 50k.}
\end{subfigure}\hfill
\begin{subfigure}{0.49\textwidth}
  \includegraphics[width=\linewidth]{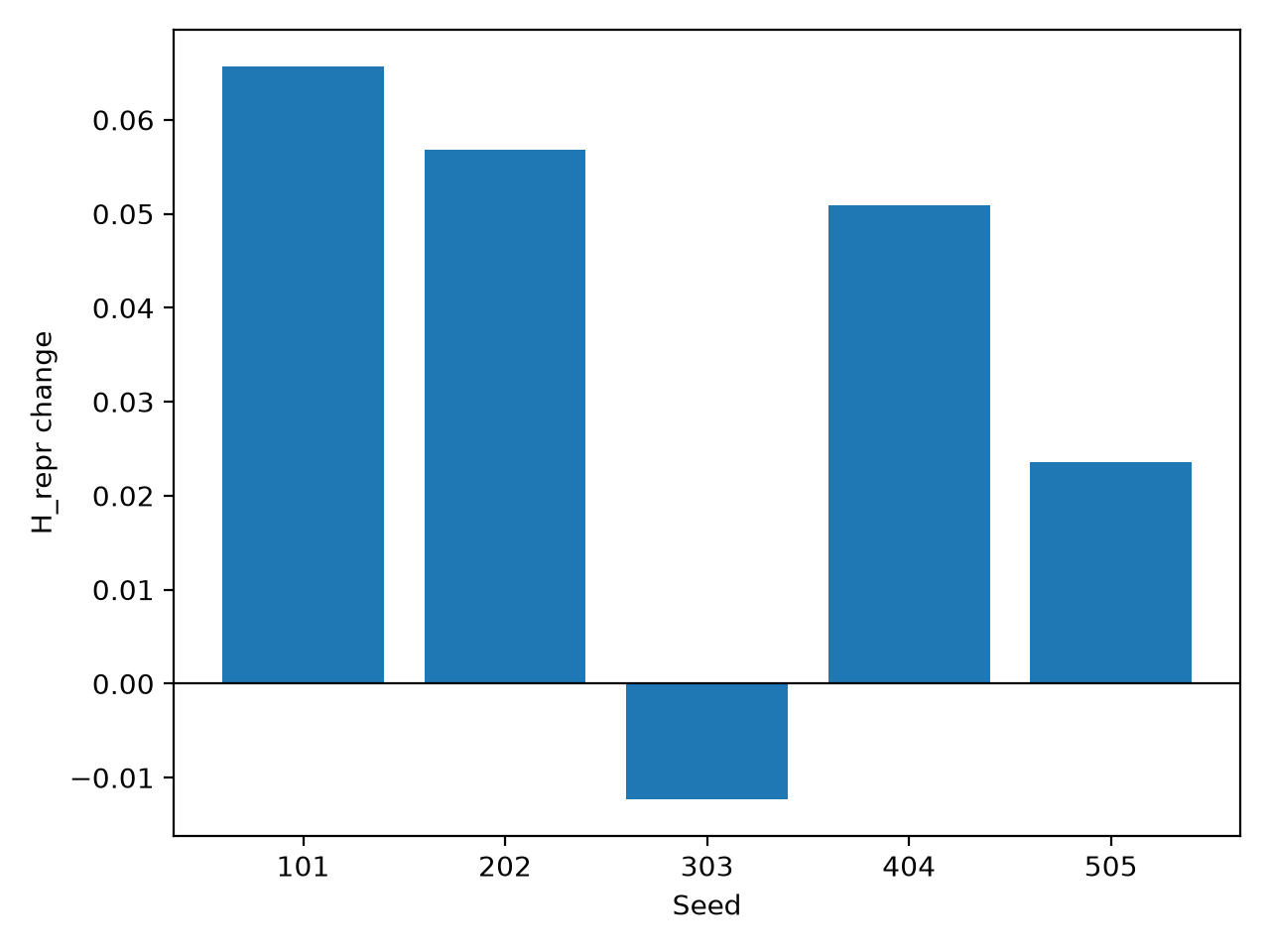}
  \caption{Per-seed change from 10k to 50k.}
\end{subfigure}
\caption{Long-horizon stress test across five paired seeds. The residue remains substantial through 50,000 common updates. Four of five seeds increase from 10k to 50k, but the sample is too small and variable to support a general growth claim.}
\label{fig:long}
\end{figure}

\subsection{Distinct representations retain similarly accessible class information at high probe-data budgets}
Frozen FC1 linear probes reveal a systematic dependence on labeled probe data. With 25 examples per class, the mean SABC-minus-SBAC probe-accuracy difference is +1.005 percentage points; the paired $t$ test gives $p=0.056$. At 50 examples per class, the difference is +0.580 points with paired $t$-test $p=0.027$, Wilcoxon $p=0.028$, and paired $d_z=0.535$. At 100 examples per class the difference falls to +0.288 points, and at 500 examples per class it is only +0.090 points.

At 500 examples per class, a TOST analysis supports practical equivalence within the predeclared $\pm0.5$ percentage-point margin: the 90\% CI for the signed difference is $[-0.037,0.217]$ percentage points and both one-sided tests reject the corresponding equivalence bounds. The 50-example effect also survives repeated fresh-head initialization: averaging over five probe seeds yields a mean SABC advantage of +0.496 percentage points, with 15 of 20 model seeds positive (paired $t$-test $p=0.011$; Wilcoxon $p=0.008$).

These results suggest that the paired representations can differ geometrically while preserving nearly equivalent task information when the readout receives sufficient supervision. Differences are more visible as sample-efficient accessibility under limited probe data.

\begin{figure}[t]
\centering
\includegraphics[width=0.73\textwidth]{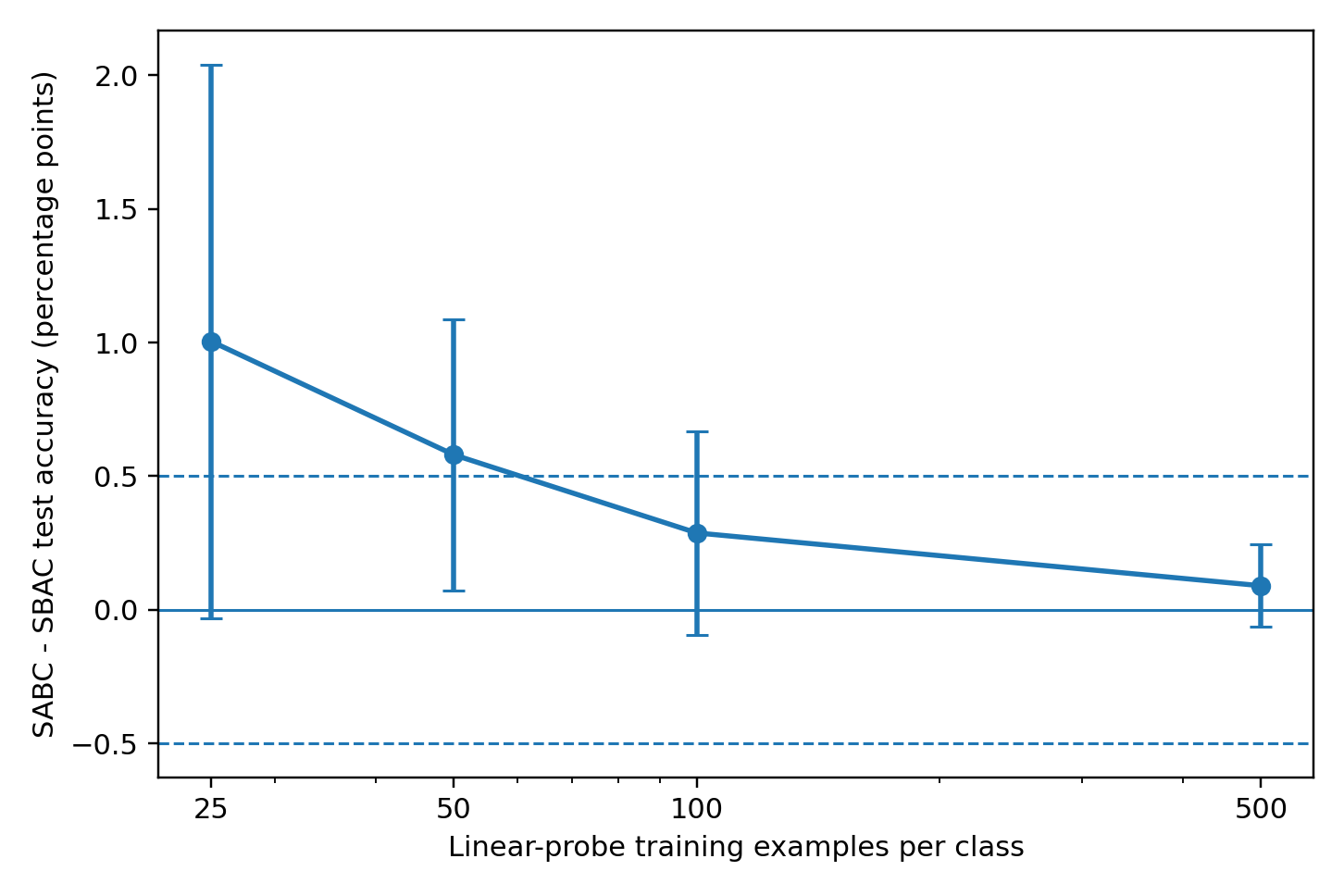}
\caption{Fresh FC1 linear-probe differences across labeled-data budgets. Points show the mean SABC-minus-SBAC full-test accuracy difference; error bars are 95\% $t$ intervals. Dashed horizontal lines mark the $\pm0.5$ percentage-point practical-equivalence margin used at 500 examples per class.}
\label{fig:probe}
\end{figure}

\begin{table}[t]
\centering
\caption{Primary FC1 linear-probe endpoint (epoch 20) across training-data budgets. Differences are SABC minus SBAC in percentage points.}
\label{tab:probe}
\small
\begin{tabular}{rrrrr}
\toprule
Examples/class & Mean diff. & 95\% $t$ CI & Paired $p$ & $d_z$ \\
\midrule
25  & +1.005 & $[-0.031, 2.041]$ & 0.056 & 0.454 \\
50  & +0.580 & $[ 0.072, 1.088]$ & 0.027 & 0.535 \\
100 & +0.288 & $[-0.094, 0.669]$ & 0.131 & 0.353 \\
500 & +0.090 & $[-0.064, 0.244]$ & 0.237 & 0.273 \\
\bottomrule
\end{tabular}
\end{table}

\subsection{LeakyReLU reduces the long-horizon residue}
At matched learning rate 0.04, ReLU and LeakyReLU have nearly identical history scores at 10,000 updates: 0.1299 and 0.1313, respectively. By 25,000 updates, the means are 0.1337 for ReLU and 0.1171 for LeakyReLU. At 50,000 updates,
\[
\hrepr^{\mathrm{ReLU}}=0.1548,\qquad \hrepr^{\mathrm{LeakyReLU}}=0.1150.
\]
The primary paired difference is therefore $\Delta\hrepr=-0.0398$. All five paired seeds are negative. The bootstrap 95\% interval is $[-0.0772,-0.0176]$, and paired effect size is $d_z=-0.97$. The paired $t$ test gives $p=0.097$ and the Wilcoxon test gives $p=0.0625$.

Functional differences decrease as well. At 50,000 updates, prediction disagreement averages 2.56\% for ReLU and 1.44\% for LeakyReLU. Mean absolute accuracy gap decreases from 0.582 to 0.130 percentage points. Given the five-seed sample and the non-significant conventional paired tests for the primary representation endpoint, we treat this as consistent directional mechanism evidence rather than definitive causal evidence.

\begin{figure}[t]
\centering
\begin{subfigure}{0.49\textwidth}
  \includegraphics[width=\linewidth]{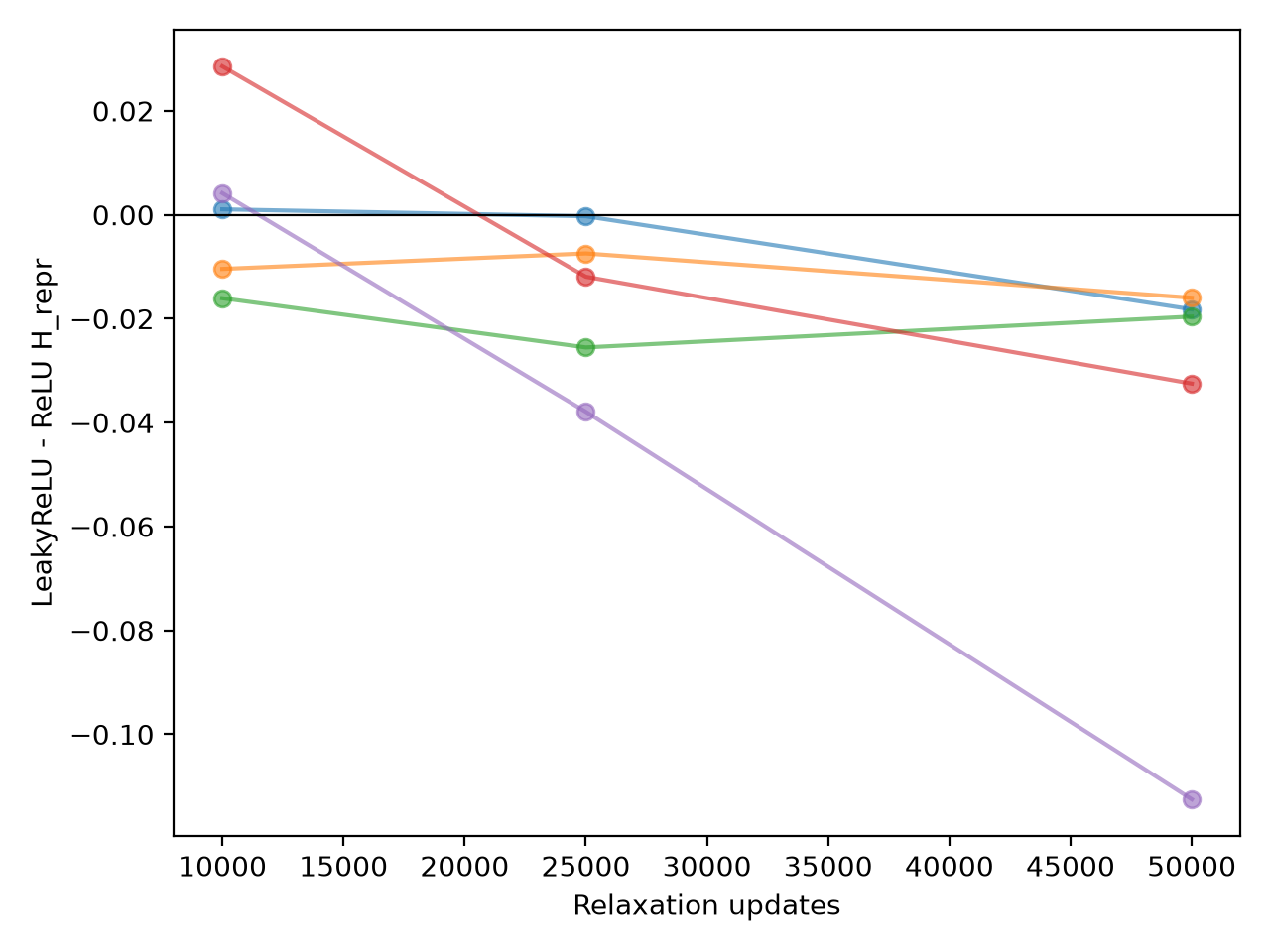}
  \caption{$\Delta\hrepr$ across the measured horizon.}
\end{subfigure}\hfill
\begin{subfigure}{0.49\textwidth}
  \includegraphics[width=\linewidth]{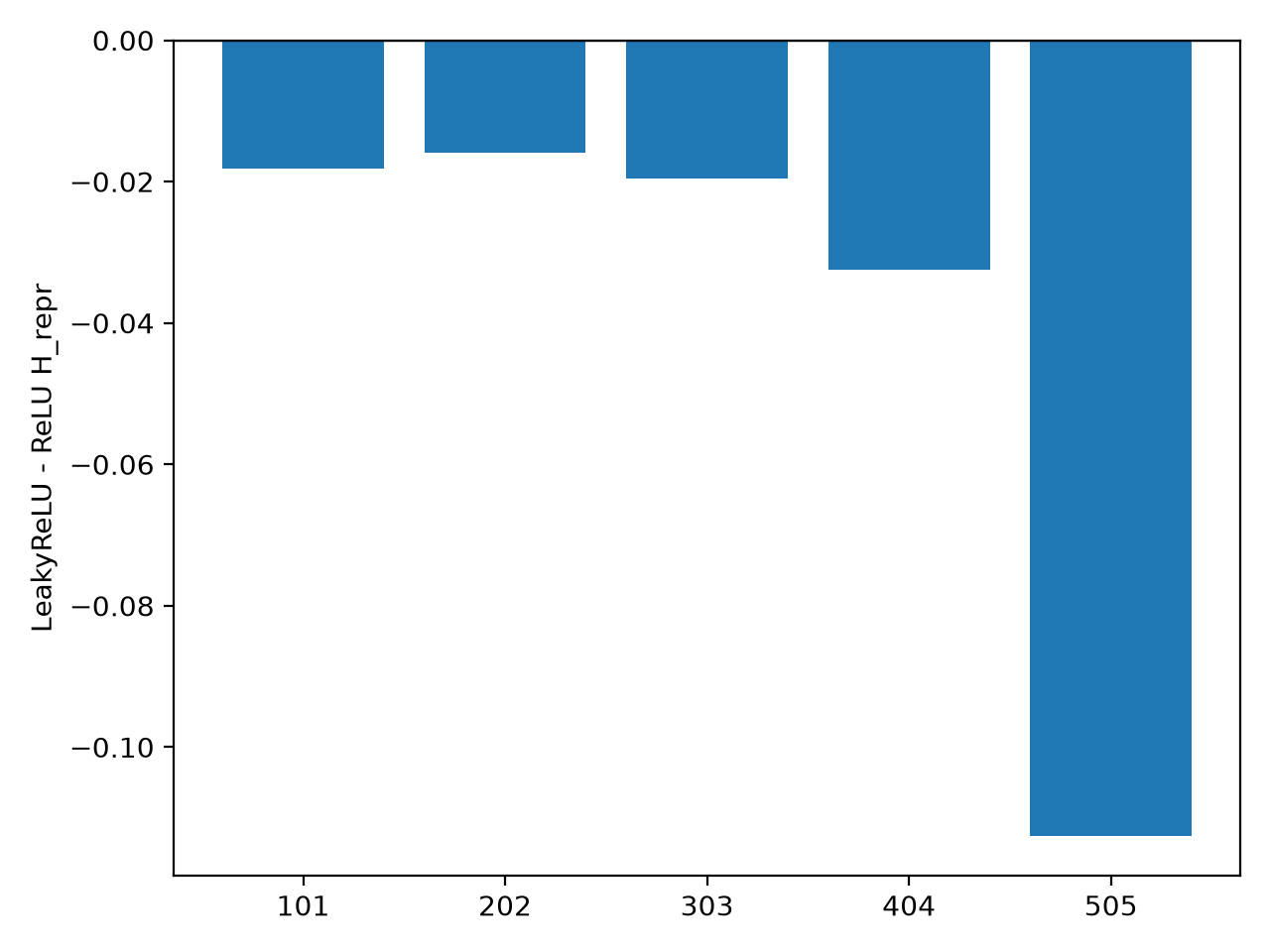}
  \caption{Per-seed $\Delta\hrepr$ at 50k.}
\end{subfigure}
\caption{Matched learning-rate activation control. $\Delta\hrepr$ is LeakyReLU minus ReLU, so negative values indicate less residual history dependence under LeakyReLU. All five seeds are negative at 50,000 updates.}
\label{fig:activation}
\end{figure}

\subsection{The effect persists when both histories use the same labels}
The rotated-MNIST experiment removes the disjoint-label structure of the main protocol. All five paired seeds satisfy the behavioral-matching criterion. At their matched endpoints, mean $\hrepr=0.1617$ with bootstrap 95\% CI $[0.1412,0.1821]$, while matched prediction disagreement averages 3.17\%.

At the common 10,000-update endpoint, mean SABC accuracy is 98.732\% and mean SBAC accuracy is 98.694\%. The mean absolute accuracy gap is only 0.112 percentage points and prediction disagreement is 2.44\%, yet $\hrepr=0.1651$ with bootstrap 95\% CI $[0.1537,0.1783]$. The persistent representation difference therefore cannot be explained solely by the original use of disjoint 0--4 and 5--9 output classes.

\begin{figure}[t]
\centering
\begin{subfigure}{0.49\textwidth}
  \includegraphics[width=\linewidth]{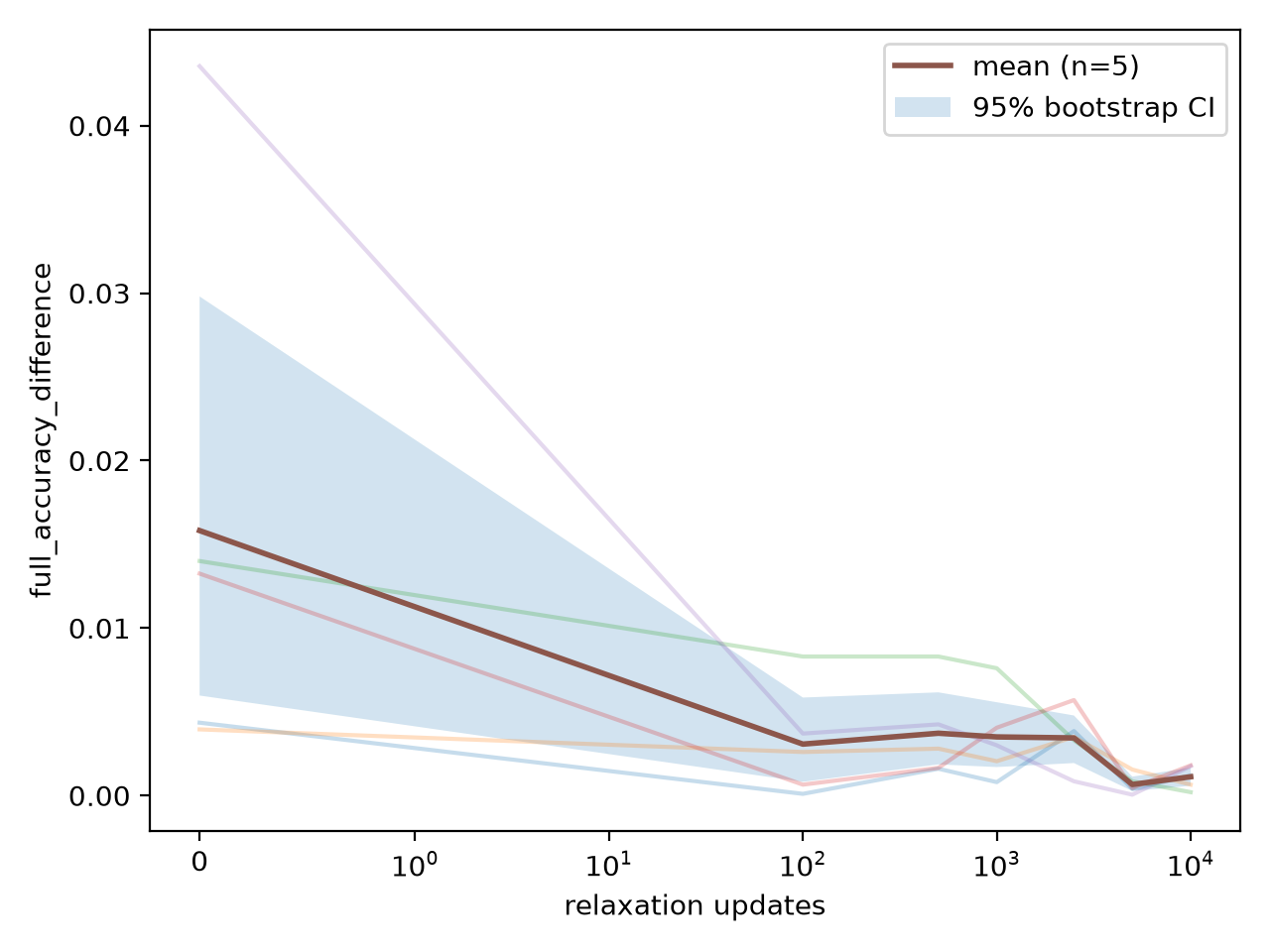}
  \caption{Accuracy-gap recovery.}
\end{subfigure}\hfill
\begin{subfigure}{0.49\textwidth}
  \includegraphics[width=\linewidth]{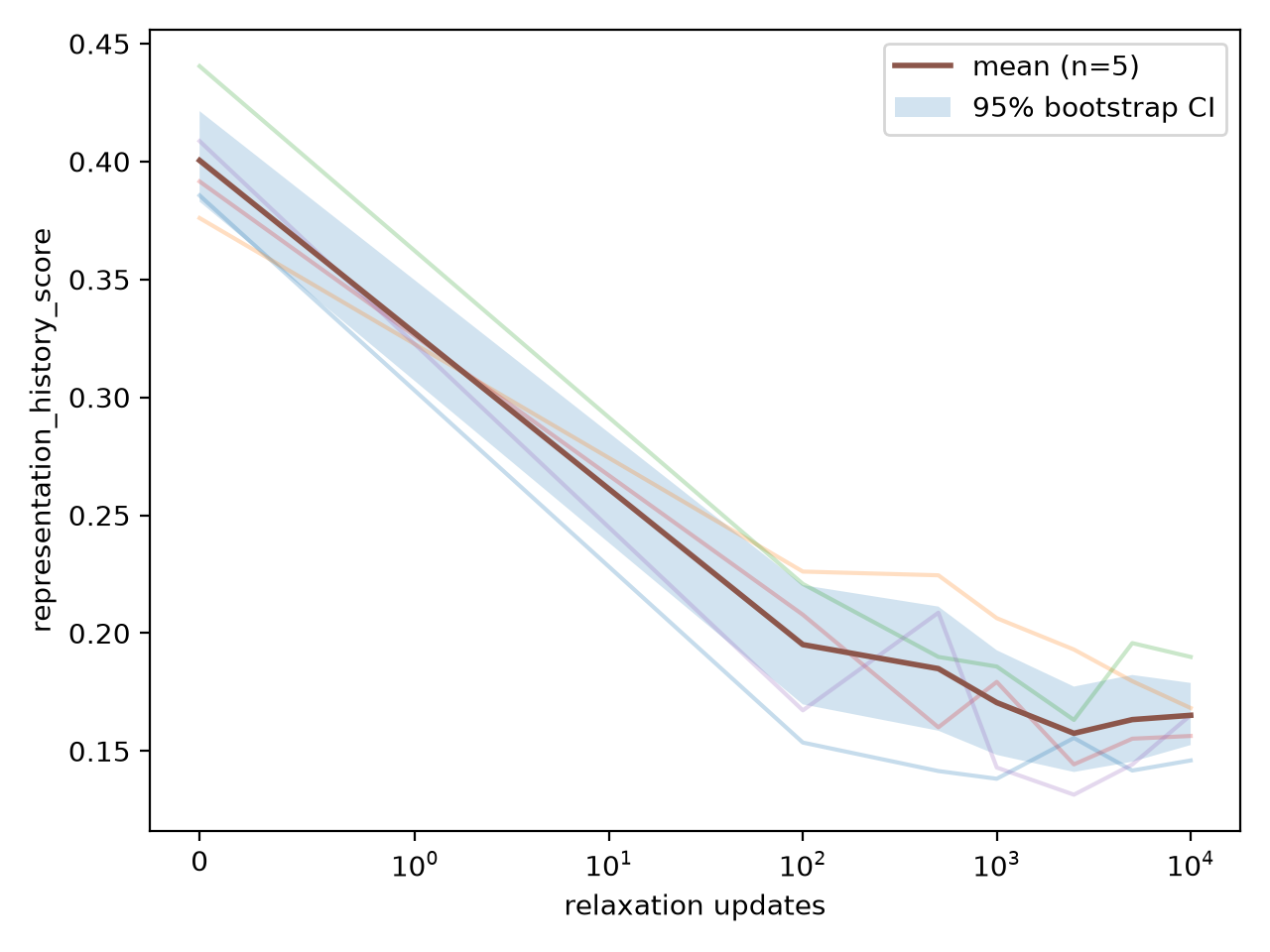}
  \caption{Representation-history score.}
\end{subfigure}
\caption{Same-label rotated-MNIST control across five paired seeds. Behavioral gaps contract while a non-zero representation residue remains.}
\label{fig:rotated}
\end{figure}

\begin{table}[t]
\centering
\caption{Summary of the main evidence chain. Bootstrap intervals are mean 95\% intervals.}
\label{tab:summary}
\small
\begin{tabular}{p{3.2cm}cccc}
\toprule
Condition & $n$ & Behavioral criterion & $\hrepr$ & 95\% bootstrap CI \\
\midrule
Main matched endpoints & 16/20 & $\leq0.2$ pp gap, $\geq97\%$ acc. & 0.139 & [0.127, 0.153] \\
Main at 10k & 20 & mean gap 0.252 pp & 0.153 & [0.144, 0.164] \\
Long horizon at 50k & 5 & mean gap 0.180 pp & 0.190 & [0.161, 0.219] \\
Rotated matched endpoints & 5/5 & same criterion & 0.162 & [0.141, 0.182] \\
Leaky minus ReLU at 50k & 5 & matched LR=0.04 & $-0.040$ & [-0.077, -0.018] \\
\bottomrule
\end{tabular}
\end{table}

\section{Discussion}

The experiments consistently separate two notions that are often treated as interchangeable: convergence in task behavior and convergence in internal representation. During common relaxation, models with different histories recover similar full-task accuracy, yet CKA continues to distinguish their deeper representations. This difference survives the mechanically defined behavioral-matching criterion, a 50,000-update stress test, and a same-label rotated-MNIST control.

One interpretation is that the final objective admits a family of functionally similar internal solutions and that optimization history influences which member of that family is reached. This is compatible with work showing that neural-network optimization can discover multiple low-loss solutions, including solutions connected by low-loss parameter-space paths \citep{draxler2018barriers,garipov2018loss}. Our experiment adds a temporal dimension: after the current training condition has become the same, the learned representation remains informative about the path preceding it.

The linear-probe results clarify what this difference does and does not mean. A CKA difference does not imply that one history has failed to learn the classification task. At high probe-data budgets, fresh classifiers recover practically equivalent performance from both frozen representations. The two models therefore retain similarly accessible label information even though that information is arranged differently internally. Under low supervision, however, the geometry appears to have modest consequences for sample-efficient accessibility.

The activation-function comparison provides an initial mechanism clue. ReLU has zero derivative for negative pre-activations, while LeakyReLU retains a non-zero slope. Replacing ReLU with LeakyReLU does not remove the history effect, but it reduces the long-horizon residue in every paired seed examined. This is consistent with the hypothesis that restricted activation-mediated plasticity contributes to persistence. It does not show that dying or low-activity ReLU units are the sole cause: LeakyReLU changes optimization dynamics throughout the model, and five seeds are insufficient to identify a unique causal mechanism.

The rotated-MNIST result is particularly important for interpretation. In the original class split, $A$ and $B$ train disjoint output classes, raising the possibility that history residue is specific to class-partitioned catastrophic forgetting. When $A$ and $B$ instead contain the same labels and differ only by a fixed domain transformation, behavioral convergence again occurs without representational convergence. This extends the observation beyond the original class-order setting.

We describe the phenomenon as \emph{persistent training-history dependence}. It is related to the intuitive notion of hysteresis because the system's measured internal state depends on its preceding path even under the same current training condition. Classical hysteresis terminology can imply stronger properties, including cyclic protocols or persistent loop structure. The present experiments do not establish those stronger claims.

\section{Limitations}

\begin{itemize}
  \item \textbf{Scale.} All experiments use MNIST and a compact CNN. The evidence supports a controlled existence result in the tested setting, not a claim that the same magnitude or mechanism must occur in large modern architectures.
  \item \textbf{Representation metric.} The primary result uses linear CKA at Conv2 and FC1. No single similarity measure uniquely characterizes internal computation, and global-average pooling removes spatial structure from convolutional features.
  \item \textbf{Behavioral criterion.} Matching is defined primarily by accuracy. Prediction disagreement, Jensen--Shannon divergence, and loss differences provide additional diagnostics, but do not exhaust every possible behavioral distinction.
  \item \textbf{Secondary-control sample size.} Long-horizon, activation, and rotated-MNIST controls use five paired seeds and therefore have lower inferential resolution than the 20-seed main experiment.
  \item \textbf{Activation-control stabilization.} Learning rate 0.04 was selected after the original LeakyReLU learning rate 0.05 produced reproducible numerical failure. The ReLU comparison was rerun at the same selected learning rate to avoid an activation--learning-rate confound, but the stability search is a pragmatic post-hoc adjustment.
  \item \textbf{Finite horizon.} Persistence through 50,000 updates is not permanence. We do not show that representations can never converge under indefinitely long common training.
  \item \textbf{Sequential rather than cyclic protocol.} A stricter hysteresis claim would benefit from a repeated cyclic schedule demonstrating reproducible loop structure under repeated traversal of a control variable.
  \item \textbf{Weight-space diagnostics.} Raw weights are not permutation aligned. Direct parameter distances and linear interpolation are therefore secondary diagnostics rather than primary evidence.
\end{itemize}

\section{Conclusion}

We studied whether neural networks that experience different training histories converge internally after they are subsequently exposed to the same final training distribution. They need not.

Across 20 paired class-split experiments, most model pairs became behaviorally matched while retaining a substantial representation-history score. The measured difference persisted through a 50,000-update common-relaxation stress test, survived a same-label rotated-MNIST control, and remained compatible with practically equivalent high-data linear-probe performance. A matched ReLU--LeakyReLU comparison reduced the residual representation difference, providing preliminary evidence that activation-mediated plasticity contributes to the effect.

The resulting picture is not one of permanently isolated solutions, nor a universal proof of neural-network hysteresis. It is a controlled demonstration that similar final behavior does not uniquely determine internal state: optimization history can remain encoded in learned representations long after its most obvious behavioral consequences have diminished.

\section*{Reproducibility Statement}
The accompanying release package contains the paper-facing aggregate analyses for the pilot, 20-seed main experiment, frozen linear probes, 50,000-update stress test, activation-function control, and rotated-MNIST control. It also contains the final-control manifests, generated configurations, orchestration scripts, an environment snapshot, and SHA256 checksums. Raw run directories remain the source of truth in the experiment repository. The final release package was separately integrity-checked before preparation of this manuscript.

\appendix
\section{Experimental Checkpoint Schedule}
The main common-relaxation checkpoints are 0, 100, 500, 1,000, 2,500, 5,000, and 10,000 optimizer updates. The long-horizon experiment additionally evaluates 25,000 and 50,000 updates. The activation comparison declares the 50,000-update $\Delta\hrepr$ endpoint as primary.

\section{Interpretation Rules Used in This Work}
Three distinctions are maintained throughout the analysis. First, behavioral matching is not treated as exact functional identity: prediction disagreement remains an explicit metric. Second, a non-zero CKA-derived history score is treated as evidence of representational difference under the declared probe, not proof of complete parameter-space basin separation. Third, persistence over a measured horizon is not described as permanence.

\end{document}